# Judge Like a Real Doctor: Dual Teacher Sample Consistency Framework for Semi-supervised Medical Image Classification

Qixiang Zhang, Yuxiang Yang, Chen Zu, Jianjia Zhang, Xi Wu, Jiliu Zhou, *Senior Member, IEEE*, Yan Wang*, *Member, IEEE*

***Abstract*—Semi-supervised learning (SSL) is a popular solution to alleviate the high annotation cost in medical image classification. As a main branch of SSL, consistency regularization engages in imposing consensus between the predictions of a single sample from different views, termed as Absolute Location consistency (AL-c). However, only AL-c may be insufficient. Just like when diagnosing a case in practice, besides the case itself, the doctor usually refers to certain related trustworthy cases to make more reliable decisions. Therefore, we argue that solely relying on AL-c may ignore the relative differences across samples, which we interpret as relative locations, and only exploit limited information from one perspective. To address this issue, we propose a Sample Consistency Mean Teacher (SCMT) which not only incorporates AL-c but also additionally enforces consistency between the samples' relative similarities to its related samples, called Relative Location consistency (RL-c). AL-c and RL-c conduct consistency regularization from two different perspectives, jointly extracting more diverse semantic information for classification. On the other hand, due to the highly similar structures in medical images, the sample distribution could be overly dense in feature space, making their relative locations susceptible to noise. To tackle this problem, we further develop a Sample Scatter Mean Teacher (SSMT) by utilizing contrastive learning to sparsify the sample distribution and obtain robust and effective relative locations. Extensive experiments on different datasets demonstrate the superiority of our method.**

## I. INTRODUCTION

Medical image classification is one of the most essential tasks in medical image analysis. Recent proposed deep learning methods have achieved remarkable success and have been widely used in diverse medical practical tasks [1-3]. However, the remarkable results highly rely on large amounts of annotated data which are difficult to acquire due to their time-consuming nature [4]. Thus, how to train a network to perform reasonably with less annotated medical data becomes a significant problem.

This work is supported by National Natural Science Foundation of China (NSFC 62371325, 62071314), Sichuan Science and Technology Program 2023YFG0263, 2023YFG0025, 2023NSFSC0497, and Opening Foundation of Agile and Intelligent Computing Key Laboratory of Sichuan Province.

Qixiang Zhang and Yuxiang Yang contribute equally to this work. Yan Wang is the corresponding author.

Qixiang Zhang, Yuxiang Yang, Jiliu Zhou and Yan Wang are with School of Computer Science, Sichuan University, Chengdu, China (e-mail: ericZhang 5915@gmail.com; yangyuxiang3@stu.scu.edu.cn; zhoujil@cuit.edu.cn; wang yanscu@hotmail.com).

Chen Zu is with Department of Risk Controlling Research, JD.com, China (e-mail: chenzu@outlook.com).

Jianjia Zhang is with School of Biomedical Engineering, Sun Yat-sen University, Shenzhen, China. (e-mail: zhangjj225@mail.sysu.edu.cn).

Xi Wu is with School of Computer Science, Chengdu University of Information Technology, China (e-mail: wuxi@cuit.edu.cn).

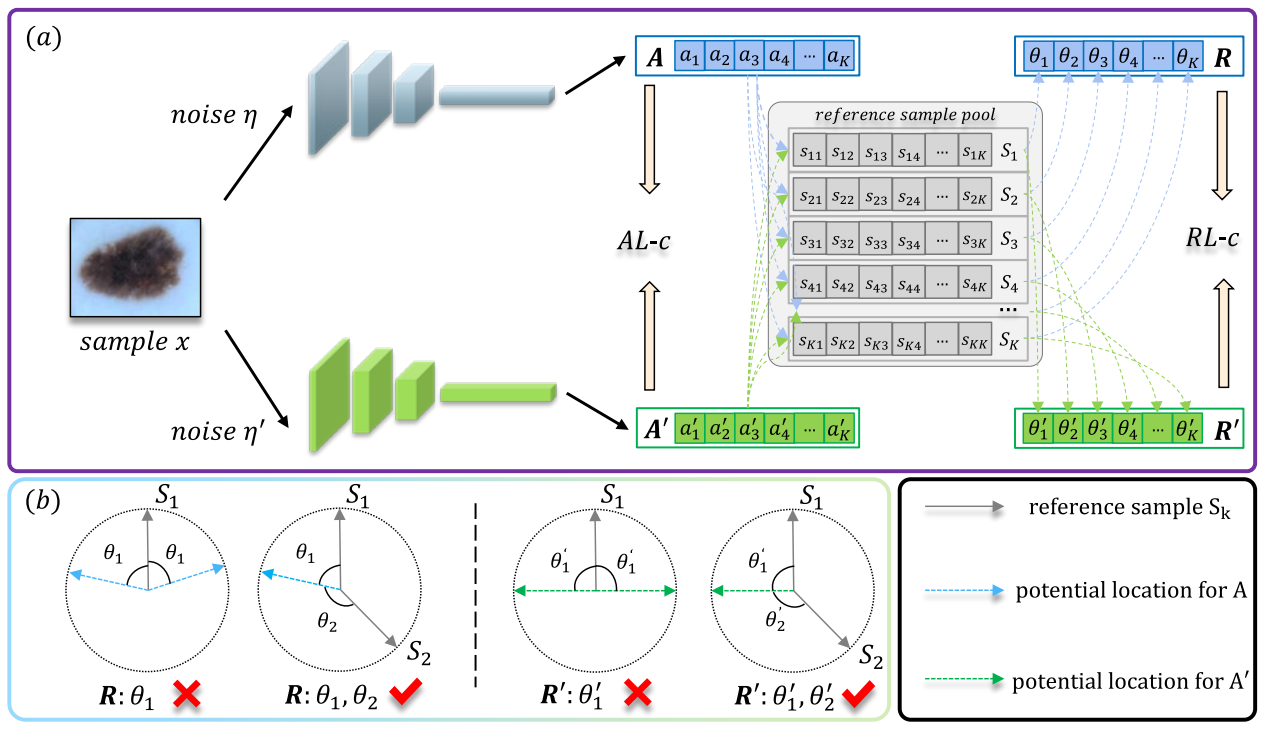

Fig. 1. The overview of Absolute Location Consistency (AL-c) and our Relative Location Consistency (RL-c). AL-c considers the prediction consistency of different views from a given sample, while RL-c analyzes the relative location consistency between different views of a given sample and those reference samples.

In response to the above issue, researchers are sparing increasing effort to semi-supervised learning (SSL) that can leverage both annotated and unannotated data. Some works have already achieved nearly tantamount performance with their supervised counterparts [5-8]. Current SSL approaches can be broadly categorized into two factions. The first one [9-12] focuses on predicting pseudo labels for unannotated data and training the model with both pseudo labels and provided annotations. In contrast, as a more mainstream approach in SSL, the second one [13, 14] follows the strategy of consistency regularization [9] which enforces the model predictions of different views of the same sample to be consistent.

Despite the achieved promising results, the above consistency-based methods only consider the individual prediction consistency of different views from a single sample, which is called Absolute Location consistency (AL-c), and suffer certain limitations. Specifically, we take a real-world clinical case for example, instead of only observing a single case, medical practitioners usually prefer to find some related trustworthy samples as references first, and then observe the relative differences between the given case and the references to determine its final diagnosis. Based on the above observation, we believe that such inter-sample differences could reflect certain clinical information for disease diagnosis.

To exploit such hidden information neglected by AL-c, in this work, we first interpret these differences as the relative location features between the given sample and the references, and then further propose a Relative Location consistency (RL-c) to constrain the relative locations of different views from a single sample to be similar. An illustration of the AL-c and the proposed RL-c is presented in Fig. 1 (a). Specifically, given an input image $x$, we can obtain its absolute location features $A$ and $A'$ with different noises under a similar design of the Mean-Teacher (MT) framework [14]. Meanwhile, we maintain a reference sample pool ($RSP$) to store the absolute location features of $K$ reference samples $S_1, \ldots S_K$. Then, we calculate the vector angle (cosine similarity) $\theta_k$ (or $\theta'_k$) between $A$ (or $A'$) and each $S_k$ to measure the difference between the given sample $x$ and each reference $S_k$. Subsequently, these angle values $\theta_k$ (or $\theta'_k$) ($k = 1, \ldots K$) are concatenated to construct the relative location feature $R$ (or $R'$). Our key insight is that if two views are from the same sample, they should be consistent in both absolute locations and relative locations. Therefore, in addition to imposing AL-c between $A$ and $A'$, we also impose RL-c between $R$ and $R'$.

However, the performance of RL-c may be hindered by two issues. Problem $\mathbb{A}$ is that due to the limitation of reference samples and angles, the relative location feature $R$ may not represent the accurate position of $x$ in the feature space. For example, assuming that there is only one reference sample $S$ in $RSP$ and one angle $\theta$ in $R$ to indicate that $x$ is positioned at an angle $\theta$ relative to $S$. In this case, $A \in \mathbb{R}^{1\times 2}$, as illustrated in Fig.1 (b), there lead to two possible locations (i.e., in the clockwise or counterclockwise direction) for $x$ around the corresponding $S_k$ in the feature space, which results in an inaccurate location of $x$. However, if two angles along with two $S_k$ are given, the final location of $A$ will be fixed. Therefore, the accuracy of $R$ requires storing enough reference samples whose number is no less than the dimension of $A$, thus deriving enough angles to constrain $A$ to a precise fixed location. Notably, in this situation, we regard $S_k$ in $RSP$ are linearly independent. Problem $\mathbb{B}$ concerns the issue of the noise induced by an overly dense sample distribution, which may readily impact the robustness of relative location feature $R$. This issue is especially pronounced in the medical imaging field, where scans targeting the same organs often exhibit highly similar features. Consequently, when samples are too close, minor noise can dramatically alter $R$, as the inter-view distances of the same sample may be comparable with the inter-sample distances. To mitigate this issue, we employ self-supervised contrastive learning [15, 16] to enhance discrimination by pulling similar (positive) pairs closer and pushing dissimilar (negative) pairs. In this work, we treat different noisy views of the same sample as positive pairs and views from different samples as negative pairs. We aim to achieve a more dispersed sample distribution and enhance the robustness of the relative location features.

Overall, in this paper, based on the MT framework and the above ideas, we present a novel Dual Teacher Sample Consistency framework, named DT-SC, where the student model is guided by a sample consistency teacher and a sample scatter teacher. Specifically, the sample consistency teacher model together with the student model forms a Sample Consistency Mean Teacher (SCMT) which encourages both RL-c and AL-c. Besides, we store more trustworthy references in a uniform reference sample pool for the student and teacher models, thus helping the DT-SC calculate the relations among samples efficiently and accurately. The sample scatter teacher model along with the student model constitutes a Sample Scatter Mean Teacher (SSMT), which ensures that the sample distribution is scattered enough when calculating relative location features. Furthermore, we employ contrastive learning to make these relations more robust. The contributions of this work can be summarized as follows:

(1) Besides the AL-c paradigm, we propose an RL-c paradigm which provides a different view on consistency regularization. By enforcing consistency on relative locations, the network can extract more diverse semantic information for classification.

(2) We introduce contrastive learning to solve the problem of overly dense sample distribution when computing relative locations. Hence, the robustness of relative location features can be improved and thus more effective feature representations can be explored.

(3) We verify our method on different datasets, and the results reveal the superior performance of our DT-SC over state-of-the-art semi-supervised learning methods.

## II. RELATED WORKS

In this section, we first review consistency-based SSL methods. Then, we discuss the self-supervised contrastive learning methods and review relevant categories of SSL methods in the medical imaging field.

### A. *Consistency-based Semi-supervised Learning*

Consistency-based SSL methods optimize the classification prediction of labeled images and minimize the prediction outputs of different views of unlabelled images, where these views are obtained from different types of image perturbations, such as spatial/temporal [13, 14], adversarial transformation [17], or data augmentation [18-20]. For example, Π-model [13] imposes consensus upon model predictions of different spatial views. The Mean Teacher (MT) framework [14] uses the student model's exponential moving average (EMA) weights to create an ensemble teacher model. This teacher model provides additional supervision, guiding the student model to progressively learn representative features from unlabeled images. Considering the data augmentation strategies, ReMixMatch [20], and FixMatch [8] applied stronger augmentations and utilized a cross-entropy loss to minimize the discrepancy between predictions. Notably, due to the ingenious design, the MT framework has become popular in consistency-based SSL methods and has achieved state-of-the-art performance on multiple practical tasks [14, 19].

### B. *Self-supervised Contrastive Learning*

Self-supervised contrastive learning aims to build a more reasonable sample distribution by performing positive pair alignment (i.e., pulling views from the same sample together) and negative pair dispersion (i.e., pushing views from different samples away) at the same time. Considering representative works of self-supervised contrastive learning, SimCLR [21] constructs both positive and negative pairs using on-the-fly batch samples. In contrast, Grill *et al.* [22] proposed BYOL, arguing against the need for negative pair dispersion, and instead implemented an asymmetric structure, bootstrapping model predictions for consistency targets. As contrastive learning and consistency-based semi-supervised learning algorithms share many similarities in methods and objectives, the performance of the consistency-based methods can be further improved with self-supervised contrastive learning [23]. For instance, Chaitanya *et al.* [24] utilize a contrastive learning loss to extract domain-specific and problem-specific cues, significantly enhancing model performance on Magnetic Resonance Imaging datasets.

In this work, we introduce the concept of contrastive learning to solve the problem of overly dense sample distribution when computing relative locations (Problem $\mathbb{B}$), aiming to achieve a more dispersed sample distribution and enhance the robustness of the relative location features.

### C. Semi-supervised Learning in Medical Imaging

To reduce the tremendous cost of manually annotated data, the community of medical imaging has also spared great efforts upon SSL that can leverage unlabeled data to assist the training in various medical image analysis tasks [7, 25, 26].

The recently proposed SSL methods used in medical imaging can be broadly categorized into four groups, including, Adversarial Learning methods, Graph-related methods, Self-supervised methods, and Consistency Regularization methods. The key insight of Adversarial Learning methods [27] is to apply generated adversarial perturbations to the input samples and force the model to be more robust to such noise. Graph-related methods aim to construct a sample graph, where each node represents a sample, and edges connect those nodes that are semantically similar. Then, a labeling strategy (e.g., label propagation [28, 29] was applied to the graph to generate pseudo-labels for those unannotated samples. Self-supervised methods mostly focus on performing pretraining with unlabeled data (not limited to medical image data) and fine-tuning afterward and have been widely used in the medical imaging community [30, 31]. Consistency Regularization Methods are built upon an assumption that predictions of different views of the same sample should be consistent. Cui *et al.* [32] first introduced the Mean Teacher (MT) framework into brain lesion segmentation. Meanwhile, Liu *et al.* [7] proposed a Sample Relation Consistency Mean Teacher (SRC-MT) network that improved the MT framework by additionally embedding a Sample Relation Consistency (SRC) paradigm that extracted semantic information from inter-sample relationships. Incorporated with the SRC paradigm, SRC-MT achieved successful improvement in skin lesion diagnosis and thorax disease diagnosis. However, it neglects the aforementioned Problem $\mathbb{A}$ and $\mathbb{B}$. Therefore, unlike previous works that only consider the AL-c paradigm, we also introduce the RL-c paradigm to provide a novel perspective on consistency regularization. This approach facilitates the network in extracting more diverse semantic information for classification, aiming to achieve the concept of "Judging Like a Real Doctor". Besides, we also incorporate contrastive learning to address the challenge of overly dense sample distributions during the computation of relative locations.

## III. METHODOLOGY

### A. Preliminary and Overall Scheme

In our problem setting, we are provided with a relatively small labeled image set $S_L = \{(x_i,\ y_i)\}_{i=1}^{N_l}$ with $N_l$ labeled images, and a larger unlabeled image set $S_U = \{x_i\}_{i=1}^{N_u}$ with $N_u$ unlabeled images, where $x_i \in \mathbb{R}^{ch\times h\times w}$ represents the $i^{th}$ image with channel number $ch$, height $h$, and width $w$, and $y_i \in \mathbb{R}^{1\times c}$ stands for the one-hot ground-truth label of $x_i$ with $c$ categories. For an image $x_i$, since its latent feature vector implies an absolute location in a multi-dimensional feature space, we define its latent feature vector as the absolute location feature. Meanwhile, $x_i$ can also be fixed in a definite location in another feature space by considering the relative difference (e.g., angle) between it and other images (called reference samples), and we define such relative difference as the relative location feature of $x_i$.

In this work, we employ the Mean Teacher (MT) framework as the backbone of our model, which aligns with [7, 14]. Fig. 2 illustrates our overall DT-SC framework, it consists of a Sample Consistency Mean Teacher (SCMT) network and a Sample Scatter Mean Teacher (SSMT) network. Both SCMT and SSMT root in the MT architecture and share the same exponential moving average (EMA) weight of the student model. Specifically, in SCMT, while maintaining the Absolute Location consistency (AL-c) on the absolute location features like the MT framework, we additionally encourage a Relative Location consistency (RL-c) to constrain the relative location features. These two consistency paradigms (i.e., AL-c and RL-c) jointly help to guide the student model to excavate semantic information from the two different location representations of the unlabeled samples from two different viewpoints. On the other hand, in SSMT, we resort to contrastive learning to enforce the views from the same sample to gather and the views from different samples to disperse, thus scattering the sample distribution. By doing this, the student model can be enhanced to produce more robust relative location features. More details of SCMT and SSMT are described in the following sub-sections.

### B. Sample Consistency Mean Teacher Network

Consistency regularization is an effective strategy to extract intrinsic information from samples to assist the learning of downstream tasks. In our SCMT, we propose two sample consistency paradigms, i.e., AL-c and RL-c. They collaboratively aid in guiding the student model to explore additional hidden information from unlabeled data, offering

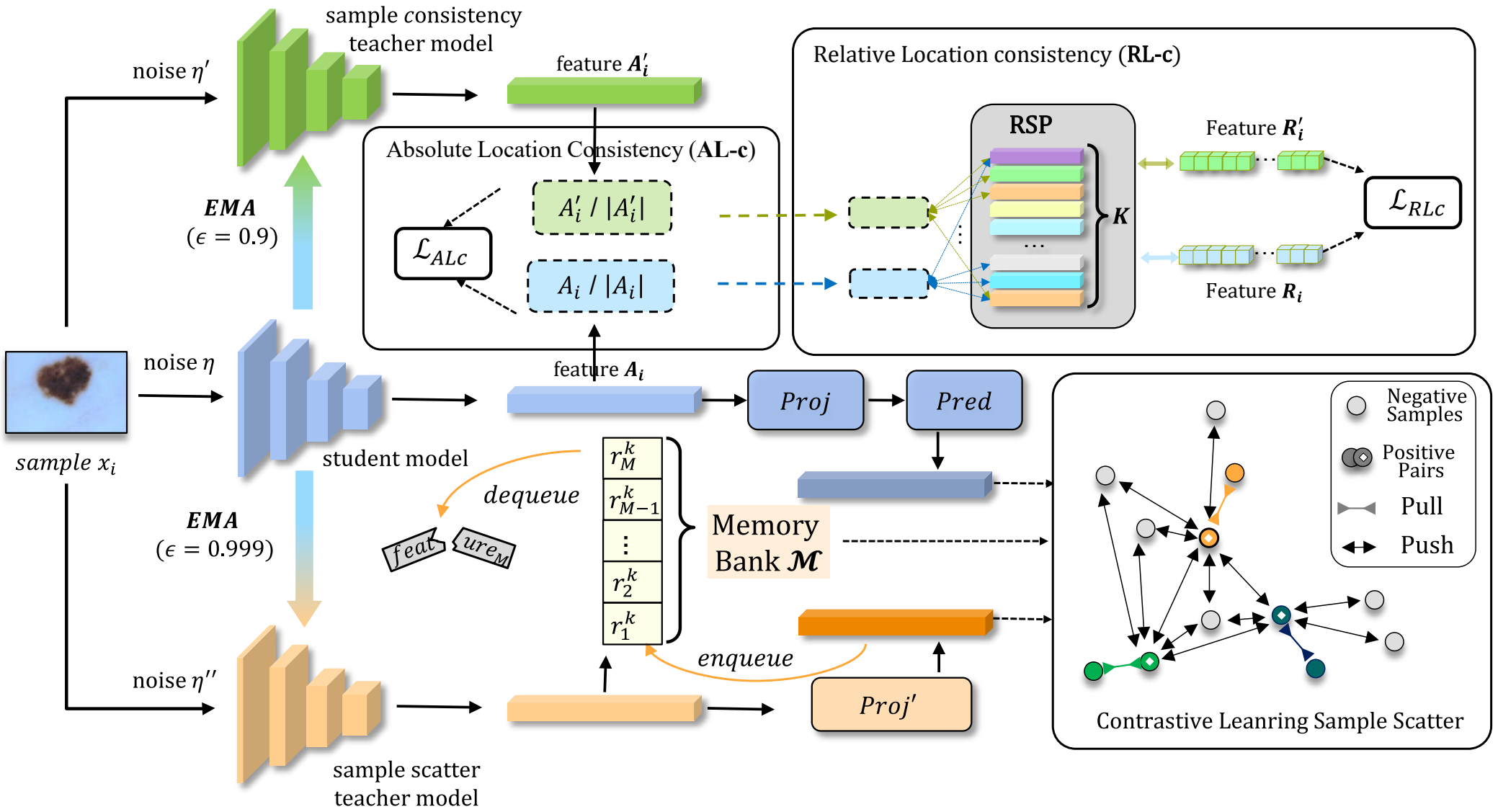

**Fig. 2.** Overview of Dual Teacher Sample Consistency framework consisting of a Sample Consistency Mean Teacher (student model + sample scatter teacher model) and a Sample Scatter Mean Teacher network (student model + sample scatter teacher model). The input sample $x_i$ is imposed with different perturbations and then fed into the different modules of the network, respectively. In SCMT, a reference sample pool $RSP$ is utilized to maintain a collection of reference samples and two consistency paradigms AL-c and RL-c collaboratively direct the student model to extract semantic information from distinct location representations of unlabeled samples. In SSMT, we utilize asymmetric structures for the student model and the teacher model by applying distinct projection operations. Additionally, a memory bank $\mathcal{M}$ is employed to retain key features from both positive and negative samples provided by the teacher model. The purpose of SSMT is to enforce the views from the same sample to gather and the views from different samples to disperse.

supplementary supervision from two distinct perspectives.

**Absolute Location consistency (AL-c).** Given a sample $x_i$, we can obtain two absolute location features $A_i \in \mathbb{R}^{1\times K}$ and $A_i' \in \mathbb{R}^{1\times K}$ as follows:

$$A_i = f(x_i, \eta, \theta), A_i' = f_{sc}(x_i, \eta', \theta_{sc}), \tag{1}$$

where $f$ and $f_{sc}$ represent the student model and the sample consistency teacher model with parameter $\theta$ and $\theta_{sc}$, $\eta$ and $\eta'$ stand for the noises imposed on $x_i$. The overall intent of AL-c is to force the student model and the teacher model to map different views of the same sample into the same location in feature space, i.e., to minimize the disparity between $A_i$ and $A_i'$. To this end, we first normalize $A_i$ and $A_i'$, and then utilize Mean Squared Error (MSE) loss on them, which can be formulated as:

$$\mathcal{L}_{ALc} = \frac{1}{N_l + N_u} \sum_{i=1}^{N_l+N_u} \left\| \frac{A_i}{|A_i|} - \frac{A_i'}{|A_i'|} \right\|_2^2, \tag{2}$$

where "$|\cdot|$" represents the L2 normalization of features.

**Relative Location consistency (RL-c).** Based on the assumption discussed in the Introduction Section, we believe that there are certain task-related clues hidden in the relative locations among various samples. Therefore, we further propose an RL-c paradigm to explore these extra clues by encouraging agreement between the relative location features of two perturbated views for a given sample. To calculate the relative location features of a given sample $x_i$, we first preserve a reference sample pool $RSP \in \mathbb{R}^{K\times K}$ to provide a certain number of reference samples. Each slot in $RSP$ stores a reference sample feature $S_i \in \mathbb{R}^{1\times K}$ ($i = 1, \ldots, K$), which is generated by mixing absolute location features:

$$S_i = \alpha \frac{A_i}{|A_i|} + (1-\alpha)\frac{A_i'}{|A_i'|}, where\ 0 < \alpha < 1. \tag{3}$$

$\alpha$ is the coefficient value to mix absolute location features, (set to 0.5). Then, regarding any reference sample feature $S_k$, we can obtain two relative location angles (cosine similarities) of $x_i$ by calculating the inner product between $A_i (or\ A_i')$ and $S_k$, as expressed below:

$$R_{i,k} = \frac{A_i}{|A_i|} \cdot (S_k)^T, R_{i,k}' = \frac{A_i'}{|A_i'|} \cdot (S_k)^T, where\ k < i. \tag{4}$$

Taking all $S_k$ into account, we finally attain the complete relative location features of two perturbated views of $x_i$:

$$R_i = [R_{i,1}, R_{i,2}, \ldots, R_{i,K}], R_i' = [R_{i,1}', R_{i,2}', \ldots, R_{i,K}']. \tag{5}$$

Then, similar to AL-c, we perform RL-c by imposing MSE loss on $R_i$ and $R_i'$ as follows:

$$\mathcal{L}_{RLc} = \frac{1}{N_l + N_u} \sum_{i=1}^{N_l+N_u} \|R_i - R_i'\|_2^2. \tag{6}$$

At the end of each epoch, the newly generated $S_i$ will be enqueued into $RSP$. Once the number of stored reference sample features reaches $K$, the oldest $S_K$ will be dequeued.

### C. *Sample Scatter Mean Teacher Network*

One key factor of the above RL-c is to derive accurate relative location features. However, when samples are distributed too dense, the generated relative location features may be overly sensitive to the noises imposed on samples, thus hampering the knowledge extraction in the inter-sample relative difference and leading to deviated relative location features. Such risk is particularly high when handling medical images due to the similar structures among different samples. To reduce such a negative effect, we propose an SSMT network to perform sample-level scatter with the help of contrastive learning [33].

The core idea of contrastive learning is to pull similar samples together and push dissimilar samples away. Consequently, it is crucial to construct paired similar samples (positive pairs) and paired dissimilar samples (negative pairs). In our setting, considering that different views of the same sample under tiny perturbations are highly semantically similar, we regard them as a positive pair. In contrast, those views from different samples are formed into negative pairs.

Concretely, with $x_i$ as an input, we can get two absolute location features $A_i$ and $A_i''$ of its two perturbated views by:

$$A_i = f(x_i, \eta, \theta);\ A_i'' = f_{ss}(x_i, \eta'', \theta_{ss}), \tag{7}$$

where $f$ and $f_{ss}$ is the student model and the sample scatter teacher model with corresponding parameter $\theta$ and $\theta_{ss}$, while $\eta$ and $\eta''$ are the perturbations imposed on $x_i$.

However, different from the SCMT network in which the student model and the teacher model are symmetrical, to prevent the student model from overly collapsing to the sample scatter teacher model during the positive pair gathering process, similar to [34], we extend the student model and the sample scatter teacher model to an asymmetric structure. Specifically, a Multilayer Perceptron (MLP) projector $Proj$ and an MLP predictor $Pred$ are additionally applied to the student model, while only one MLP projector $Proj'$ is attached to the sample scatter teacher model. Then, we can obtain the query feature $r_i^q$ and the key feature $r_i^k$ by:

$$r_i^q = Pred\big(Proj(A_i)\big), r_i^k = Proj'(A_i''). \tag{8}$$

It is worth noting that we maintain a progressively updated memory bank $\mathcal{M}$ with M slots to store key features $\mathcal{M} = [r_1^k, r_2^k, \dots, r_M^k]$. The on-the-fly query $r_i^q$ and its corresponding key $r_i^k$ from one $x_i$ form a positive pair and the query $r_i^q$ along with $r_j^k$s $(j < i)$ in $\mathcal{M}$ produce negative pairs. Following [15], we utilize InfoNCE [6] to build the contrastive loss $\mathcal{L}_{CL}$ as:

$$\mathcal{L}_{CL} = \sum_{i=1}^{N_l+N_u} -\log \exp\left(\frac{r_i^q \cdot r_i^k}{\tau}\right) \Big/ \sum_{j=1}^{M} \exp\left(\frac{r_i^q \cdot r_j^k}{\tau}\right), \tag{9}$$

where $\tau$ is a temperature hyper-parameter (set to 0.7 as [41]).

### D. Training Process

We divide the whole training process into two phases: Sample Consistency Phase and Sample Scatter Phase. To reduce the negative effect of overly dense sample distribution, we go through the Sample Scatter Phase first to make sure the sample distribution is sparse enough before RL-c is involved.

Sample Scatter Phase. In this phase ($0 \leq e \leq \text{epoch}_{ss}$), only the SSMT network is utilized. We train the student model with the supervised loss on $S_L$ and the contrastive loss $\mathcal{L}_{CL}$ on $S_L$ and $S_U$. The loss function of this phase is termed as $\mathcal{L}_{SS}$:

$$\mathcal{L}_{SS} = \mathcal{L}_s + \lambda_{CL}\mathcal{L}_{CL}, \tag{10}$$

where $\mathcal{L}_s$ leverages the cross-entropy loss and $\lambda_{CL}$ is the coefficient for balancing these two terms.

**Algorithm I** Pseudocode of training DT-SC framework

**Input**: Labeled data $S_L = \{(x_i,\ y_i)\}_{i=1}^{N_l}$, unlabeled data $S_U = \{x_i\}_{i=1}^{N_u}$, Training epoch $E$, Sample scatter epoch $epoch_{SS}$.

**Initialize**: Student model $\{f, \theta\}$; sample consistency teacher model $\{f_{sc}, \theta_{sc}\}$, sample scatter teacher model $\{f_{ss}, \theta_{ss}\}$; Reference sample pool $RSP$; Memory bank $\mathcal{M}$.

```
for e = 0 → E do
  for i = 0 → N_l + N_u do
    if x_i ∈ S_L then
      compute L_s using cross entropy loss function.
    end if
    if e ≥ epoch_SS then
      generate A_i, A'_i based on (1);
      generate R_i, R'_i based on (4) and (5);
      compute L_ALc, L_RLc, L_SC based on (2), (6) and (11);
      update θ with regard to L_SC;
    else
      generate r_i^q, r_i^k based on (7) and (8);
      compute L_CL, L_SS based on (9) and (10);
      update θ with regard to L_SS;
    end if
  enqueue and dequeue to M and RSP.
  update θ_sc, θ_ss as EMA of θ with ϵ = 0.9 and 0.999
```

Sample Scatter Phase. In this phase ($epoch_{ss} \leq e \leq E$), we abandoned the SSMT network and began to exploit the SCMT network to extract semantic information from absolute location features and relative location features. The total loss for this phase is termed as $\mathcal{L}_{SC}$, which can be expressed as:

$$\mathcal{L}_{SC} = \mathcal{L}_s + \lambda_A\mathcal{L}_{ALc} + \lambda_R\mathcal{L}_{RLc}, \tag{11}$$

where $\lambda_A$ and $\lambda_R$ are tradeoff hyper-parameters to balance the contribution of supervised loss $\mathcal{L}_s$, $\mathcal{L}_{ALc}$ and $\mathcal{L}_{RLc}$. For better understanding, we present the training pipeline in **Algorithm I**.

## IV. Experiments and Results

### A. Dataset and Evaluation Metrics

**Pneumonia Chest X-ray dataset** [34]. This dataset comprises 5856 chest X-ray images across three categories: Bacterial Pneumonia, Virus Pneumonia, and Normal. It allocates 5233 images (90%) for training, 423 images (7%) for testing, and 200 images (3%) for validation.

**ISIC challenge 2019 dataset** [35]. This dataset is a long-tailed dataset containing 25331 dermoscopy images of skin lesions categorized into 8 types: Melanoma (MEL), Melanocytic nevus (NV), Basal cell carcinoma (BCC), Actinic keratosis (AK), Benign keratosis (BKL), Dermatofibroma (DF), Vascular lesion (VASC), and Squamous cell carcinoma (SCC). The proportion and amount of samples of each disease type are summarized in Table I. We distribute 17731 (70%) of the samples to the training set, 5066 (20%) to the testing set, and 2534 (10%) to the validating set.

TABLE I
DATA STATISTICS OF ISIC CHALLENGE 2019 DATASET.

| | MEL | NV | BCC | AK | BKL | DF | VASC | SCC | Total |
|---|---|---|---|---|---|---|---|---|---|
| Number | 4522 | 12857 | 3322 | **867** | 2624 | **239** | **253** | **628** | 25331 |
| Percent | 17.9% | 50.8% | 13.1% | **3.4%** | 10.4% | **0.9%** | **0.9%** | **2.5%** | 100% |
| Image Samples | 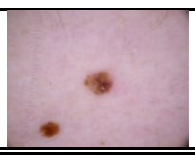 | 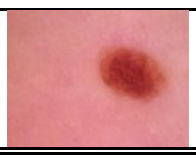 | 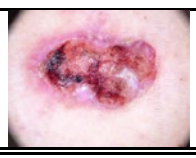 | 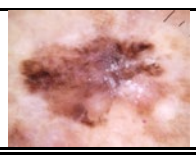 | 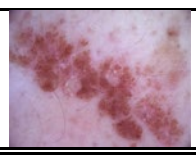 | 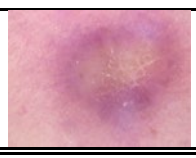 | 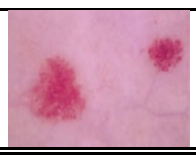 | 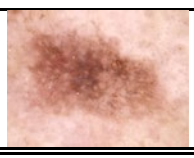 | |

TABLE II
COMPARISON RESULTS ON PNEUMONIA CHEST X-RAY DATASET. **BOLD** AND UNDERLINES DENOTE BEST AND SECOND-BEST RESULTS. NOTE THAT THE PAIRED T-TESTS BETWEEN ARE CONDUCTED ON ACCURACY AND F1 VALUE OF TEN-FOLD VALIDATION EXPERIMENTS. † DENOTES OUR REPRODUCED RESULTS.

| Method | $N_l$ | $N_u$ | Acc (%) | p-val | AUC (%) | Precision (%) | Recall (%) | F1 (%) | p-val |
|---|---|---|---|---|---|---|---|---|---|
| Upper Bound | 100% | 0 | 92.27±0.49 | | 97.80±0.33 | 91.63±0.50 | 91.63±0.28 | 91.52±0.30 | |
| Baseline | 20% | 0 | 87.66±1.02 | <0.001 | 95.33±1.37 | 86.54±0.93 | 86.24±1.22 | 86.52±0.95 | <0.001 |
| Pseudo-label [10] | 20% | 80% | 88.89±2.20 | 0.021 | 96.15±1.73 | 88.15±1.57 | 88.08±1.99 | 87.99±2.05 | 0.013 |
| MixMatch [18] | 20% | 80% | 89.58±1.77 | 0.043 | 96.20±1.62 | 88.24±0.98 | 89.75±0.48 | 88.83±0.58 | 0.036 |
| FixMatch [8] | 20% | 80% | 90.06±1.49 | 0.049 | 96.12±1.44 | 89.17±1.72 | 89.34±1.53 | 89.25±1.44 | 0.033 |
| CoMatch [6] | 20% | 80% | 89.94±2.02 | 0.042 | 96.16±2.50 | 89.24±2.33 | 89.12±2.46 | 89.14±1.93 | 0.042 |
| SimPLE [5] | 20% | 80% | 90.02±1.20 | 0.047 | 97.36±1.35 | 88.22±1.57 | 88.38±1.09 | 88.53±1.33 | 0.029 |
| MT [14] | 20% | 80% | 89.75±0.79 | 0.035 | 96.56±0.63 | 88.39±1.22 | 88.82±0.58 | 88.29±0.60 | 0.025 |
| SRC-MT [7] | 20% | 80% | 89.10±0.29 | 0.022 | 96.99±0.55 | 88.02±0.38 | 88.55±0.66 | 88.05±0.51 | 0.031 |
| CRCKD [42] | 20% | 80% | 90.33±1.27 | <0.001 | <u>97.66±0.94</u> | 89.33±1.07 | 89.73±1.18 | 89.58±1.30 | <0.001 |
| SSCL [39] | 20% | 80% | 88.29±1.43 | <0.001 | 95.57±0.95 | 87.57±1.05 | 88.60±1.72 | 87.69±0.99 | <0.001 |
| ACPL [44] | 20% | 80% | <u>90.42±2.00</u> | <0.001 | 96.77±2.35 | 89.00±1.66 | 88.54±1.93 | <u>89.77±1.96</u> | <0.001 |
| RAC-MT [40] | 20% | 80% | 89.57±1.14 | <0.001 | 97.57±0.74 | <u>89.46±1.26</u> | 90.29±1.55 | 89.69±1.25 | <0.001 |
| UniMatch [37]† | 20% | 80% | 89.92±0.97 | <0.001 | 97.39±1.07 | 88.76±1.42 | **90.59±0.88** | 89.49±1.41 | <0.001 |
| AugSeg [38]† | 20% | 80% | 88.57±0.68 | 0.022 | 96.58±0.61 | 87.94±0.73 | 89.22±1.24 | 88.93±0.92 | 0.030 |
| SemiCVT [41]† | 20% | 80% | 90.36±2.30 | <0.001 | 97.00±1.80 | 88.95±2.27 | 88.03±2.56 | 88.26±2.40 | <0.001 |
| ESL [43]† | 20% | 80% | 88.55±1.09 | 0.039 | 96.30±1.33 | 87.44±1.84 | 88.23±1.66 | 87.66±1.28 | 0.027 |
| **Proposed** | 20% | 80% | **90.58±0.32** | | **97.74±0.88** | **89.73±0.49** | <u>89.87±0.74</u> | **89.80±0.59** | |

**Alzheimer's dataset** [36]**.** This dataset consists of 6400 MRI images collected from various websites with each label verified. The images are classified into four classes according to the stage of the disease: Mild Demented (1299), Moderate Demented (551), Non Demented (2759), and Very Mild Demented (1791). The training set contains 4,098 images, with 1,023 images used for validation. The test set contains 1,279 images.

**Evaluation Metrics:** To fully evaluate our proposed DT-SC framework, we choose five different evaluation metrics, i.e., Accuracy, AUC, Precision, Recall, and F1 value on the Pneumonia Chest X-ray dataset. For the other two datasets, we adopt the average AUC and AUC of each disease type to further demonstrate the effectiveness of DT-SC in discriminating categories that own fewer samples.

### *B. Compared methods*

To demonstrate the superiority of our proposed method DT-SC, we compare it with multiple existing semi-supervised methods, including consistency regularization methods, e.g., Mean Teacher (MT) [14], and Sample Relation Consistency Mean Teacher (SRC-MT) [7], UniMatch [37], AugSeg [38], SSCL [39], RAC-MT [40] and pseudo labeling methods, e.g., Pseudo-label [10], MixMatch [19], FixMatch [8], CoMatch [6], SimPLE [5], SemiCVT [41], CRCKD [42], ESL [43], and ACPL [44]. The Baseline and Upper Bound methods are the backbone network trained in a supervised manner. To ensure a fair comparison, the results of these methods are obtained either from their respective papers or by reimplemented using their released code. In our experiments, we maintain a fixed random seed over 5 runs and report the mean results. We maintain consistency by employing the same backbone architecture and data preprocessing routines for all compared methods on each dataset.

### *C. Implementation Details*

The experiments are implemented by Pytorch 1.7 and conducted through two NVIDIA RTX 3090 GPUs. On the Pneumonia Chest X-ray dataset, we employ Convolution Vision Transformer-13 (CvT-13) [41] and resize the image size to 256×256. On the ISIC challenge 2019 dataset, we follow SRC-MT [7]. We employ DenseNet-121 [46] and resize the image size to 224×224. On the Alzheimer's Dataset, we also employ DenseNet-121 and follow [36] to operate the preprocessing procedure. For the updating strategy, our SCMT and SSMT networks typically follow the MT framework [14]. Notably, we set the coefficient of the exponential moving average as 0.999 in the SSMT network to achieve a slowly updating teacher [16]. We use Adam optimizer to train the model for 100 epochs with a learning rate of 1e-4 and batch size of 64 (16 for labeled data, 48 for unlabeled data). In the Sample Scatter Phase, the SSMT network is trained for 20 epochs, while in the Sample Consistency Phase, the SCMT network is trained for 80 epochs. Notably, to minimize memory usage, we match the reference sample pool ($RSP$) size to the absolute location features' dimension, both set at 1024. The size M of the memory bank $\mathcal{M}$ is set to 4096. The hyper-parameters are selected based on our ablation studies. Concretely, we set $\lambda_{CL}$ to 0.1 in Eq. (10), and set $\lambda_A$, $\lambda_R$ to 1 in Eq. (11). The analysis of framework designs, as well as the sensitivity test experiment for the hyperparameters, will be discussed in the subsequent analytical experiments.

TABLE III
COMPARISON RESULTS ON ISIC CHALLENGE 2019 DATASET WITH AUCS OF EIGHT DISEASE TYPES. **BOLD** AND UNDERLINES DENOTE BEST AND SECOND-BEST RESULTS.

| Method | Upper Bound | Baseline | MT [14] | SRC-MT [7] | FixMatch [8] | CoMatch [6] | RAC-MT [40] | **Proposed** |
|---|---|---|---|---|---|---|---|---|
| Proportion | $N_l$ $N_u$<br>100 0 | $N_l$ $N_u$<br>10 0 | $N_l$ $N_u$<br>10 90 | $N_l$ $N_u$<br>10 90 | $N_l$ $N_u$<br>10 90 | $N_l$ $N_u$<br>10 90 | $N_l$ $N_u$<br>10 90 | $N_l$ $N_u$<br>10 90 |
| **Average** | 94.01±3.72 | 86.82±5.15 | 88.60±3.92 | 88.79±0.49 | 89.53±4.21 | 89.15±3.36 | 90.52±3.86 | **90.56**±3.84 |
| MEL | 91.37 | 83.60 | 84.99 | 82.26 | 83.59 | **85.02** | 84.33 | 84.66 |
| NV | 94.75 | 90.19 | 91.67 | 91.63 | 91.99 | 90.35 | **92.02** | 90.77 |
| BCC | 95.58 | 94.03 | 95.00 | 94.54 | 95.16 | 94.25 | 94.99 | **95.93** |
| AK | 89.71 | 83.07 | 89.24 | 89.31 | 89.71 | 89.11 | 89.94 | **90.53** |
| BKL | 90.72 | 82.83 | 84.72 | 84.67 | 84.99 | 83.29 | 84.69 | **85.76** |
| DF | 99.07 | 88.05 | 84.88 | 87.10 | 88.26 | 91.67 | **93.44** | 92.93 |
| VASC | 100.0 | 93.70 | 92.93 | 93.20 | 94.18 | 91.27 | 94.47 | **95.28** |
| SCC | 90.87 | 79.11 | 85.33 | 84.34 | 88.35 | 88.21 | **90.33** | 88.75 |

TABLE IV
COMPARISON RESULTS ON ALZHEIMER'S DIAGNOSIS DATASET WITH AUCS OF FOUR DISEASE TYPES (MILD DEMENTED, MODERATE DEMENTED, NON DEMENTED, AND VERY MILD DEMENTED). BOLD AND UNDERLINE DENOTE BEST AND SECOND-BEST RESULTS.

| Method | Upper Bound | Baseline | MT [14] | SRC-MT[7] | FixMatch [8] | CoMatch [6] | RAC-MT[40] | **Proposed** |
|---|---|---|---|---|---|---|---|---|
| Proportion | $N_l$ $N_u$<br>100 0 | $N_l$ $N_u$<br>20 0 | $N_l$ $N_u$<br>20 80 | $N_l$ $N_u$<br>20 80 | $N_l$ $N_u$<br>20 80 | $N_l$ $N_u$<br>20 80 | $N_l$ $N_u$<br>20 80 | $N_l$ $N_u$<br>20 80 |
| **Average** | 87.72±4.22 | 82.07±5.40 | 82.29±3.41 | 85.24±4.11 | 85.17±3.76 | 84.35±2.68 | 88.33±3.95 | **89.13±4. 06** |
| Mild Dem. | 91.37 | 83.60 | 84.99 | 82.26 | 83.59 | **85.02** | 84.33 | 84.66 |
| Moderate Dem. | 99.07 | 88.05 | 84.88 | 87.10 | 88.26 | 91.67 | **93.44** | 92.93 |
| Non Dem. | 100.0 | 93.70 | 92.93 | 93.20 | 94.18 | 91.27 | 94.47 | **95.28** |
| Very Mild Dem. | 90.87 | 79.11 | 85.33 | 84.34 | 88.35 | 88.21 | **90.33** | 88.75 |

## D. *Comparison with the State-of-the-art methods*

**Comparison Results on Pneumonia Chest X-ray Dataset**. As shown in Table II, our DT-SC framework outperforms all other methods with the tall evaluation metrics. Specifically, compared with the fully supervised method (Baseline), our DT-SC can leverage unlabeled data and significantly improve Accuracy (+2.92%), AUC (+2.41%), and F1 value (+3.28%). Unlike the previous semi-supervised method SRC-MT, which extracts information based on inter-sample relations but neglects issues identified in Problem $\mathbb{A}$ and $\mathbb{B}$, our DT-SC presents a more robust alternative. By utilizing the SCMT and SSMT networks to perform multi-perspective consistency regularization and contrastive learning, DT-SC can extract more extra semantic information from unlabeled data and push views across samples away. Therefore, our DT-SC solves Problem $\mathbb{A}$ and $\mathbb{B}$ efficiently, achieving higher performance in Accuracy (+1.48%), AUC (+0.75%), and F1 value (+1.75%). Moreover, compared with the second-best method ACPL our method has demonstrated remarkable gains in AUC (+0.97%), Precision (0.97%), and Recall (1.33%). Notably, though the absolute improvements in ACC and F1 over the ACPL are modest at 0.16% and 0.03% respectively, these increments are still significant as even minor enhancements can lead to substantial improvements in clinical diagnostics. Moreover, our method exhibits lower variance in results, indicating greater stability compared to ACPL. To check whether the aforementioned improvements are statistically significant, we conduct paired t-tests on Accuracy and F1 value of ten-fold cross-validation experiments. As seen from Table II, our proposed method has statistically significant improvements over all of the competitive methods with all p-val<0.05.

**Comparison Results on ISIC Challenge 2019 dataset.** As shown in Table III, in comparison to the fully-supervised method (Baseline), our DT-SC effectively harnesses the additional unlabeled data, resulting in a notable increase in the average AUC from 86.82% to 90.56%. Please note that our method especially achieves giant improvement in the AUCs of those categories that possess less than 5% of samples, i.e., AK (+7.46%), DF (+4.88%), VASC (+1.58%) and SCC (+9.64%). The possible reason is that the RL-c paradigm involves inter-sample relations between low-profile categories and high-profile categories. Thus, the samples of high-profile categories provide extra semantic information to assist the diagnosis of the low-profile categories. In terms of semi-supervised methods, our proposed method outperforms all of the chosen SOTAs in the average AUC with relatively lower variance and also achieves top-level AUCs of most disease types. Furthermore, we also conduct experiments to compare DT-SC, SRC-MT, and RAC-MT, all of which try to enhance MT by involving inter-sample relations. As shown in Table III, all methods improve the performance of MT. However, our method DT-SC stands out by applying the Rc-L paradigm to enforce relative location consistency and utilizing contrastive learning to enhance the dispersion of the sample distribution, thus our DT-SC can enhance the performance of MT most. Besides, our DT-SC outperforms RAC-MT in many categories, i.e., AK (+0.59%), BKL (+1.07%), and VASC (+0.81%).

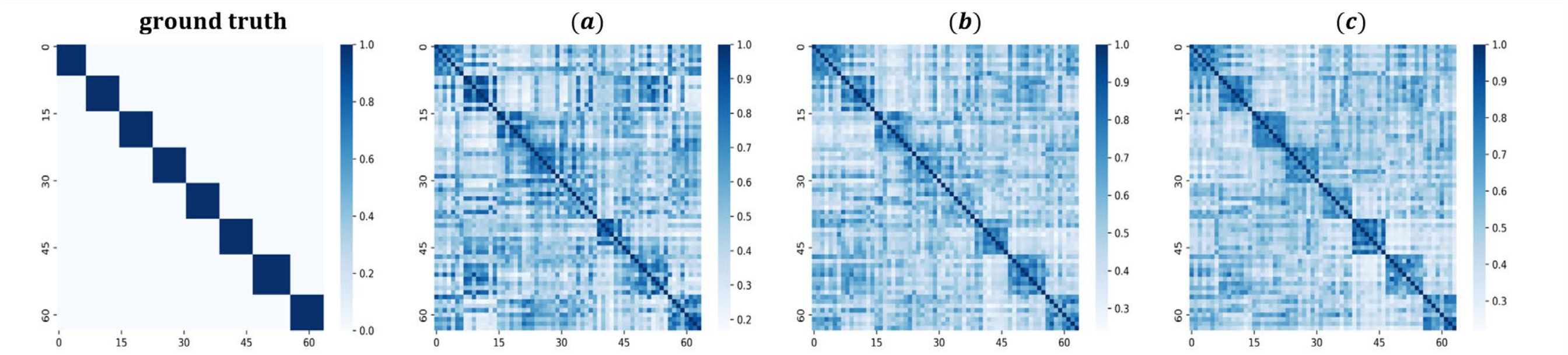

**Fig. 3**. Visualization of the similarity matrices between batch (size = 64) and some of reference samples (amount = 64) in RSP after 100 epochs on ISIC challenge 2019 dataset: (a) Pure MT framework; (b) SCMT network; (c) DT-SC framework. The major improvements are marked by red boxes.

TABLE V
ABLATION STUDIES ON PNEUMONIA CHEST XRAY DATASET. **BOLD** AND UNDERLINES DENOTE BEST AND SECOND-BEST RESULTS.

| Method | Accuracy | AUC | F1 |
|---|---|---|---|
| Upper Bound | 92.31% | 97.82% | 91.55% |
| Baseline | 87.66% | 95.46% | 86.59% |
| Pure MT | 89.26% | 96.87% | 88.05% |
| SCMT | 90.27% | 97.28% | 89.39% |
| DT-SC | **90.58%** | **97.74%** | **89.80%** |

**Comparison Results on Alzheimer's dataset**. We verify our DT-SC framework on a 3D brain MRI Alzheimer's dataset. As observed from Table IV, due to the remarkable consistency regularization ability and the utilization of contrastive learning effectively address the problem of overly dense sample distributions, our DT-SC framework achieves an average performance of 89.13% AUC, surpassing all benchmarked methods in three out of four categories: Mild Demented, Moderate Demented, and Very Mild Demented. Furthermore, our DT-SC framework not only outperforms the second-best method, RAC-MT, by a significant 0.8% margin in the AUC metric, but it also impressively surpasses the performance of the fully supervised upper bound. Notably, despite our method having a relatively higher variance, this variance does not detract from its effectiveness, as it still surpasses many competitors like SRC-MT, FixMatch, CoMatch, and RAC-MT, which are more consistent but less effective.

### *E. Ablation Study*

To give a more detailed analysis of our method, we conducted several analytical studies. All of the variant models are trained with the same training setting as described above.

**Ablation Studies on SCMT Network.** The SCMT network is designed to extract extra semantic information from unlabeled data through performing multi-perspective consistency regularization, i.e., AL-c and RL-c paradigms.

TABLE VI
ABLATION STUDIES ON ISIC CHALLENGE 2019 DATASET WITH AUCS OF 8 DISEASE TYPES. **BOLD** AND UNDERLINES DENOTE BEST AND SECOND-BEST RESULTS.

| Method | Upper Bound | Baseline | Pure MT | SCMT | DT-SC |
|---|---|---|---|---|---|
| **Average AUC** | 94.01% | 86.82% | 88.60% | 90.46% | **90.56%** |
| MEL | 91.37% | 83.60% | **84.99%** | 84.71% | 84.66% |
| NV | 94.75% | 90.19% | **91.67%** | 91.60% | 90.77% |
| BCC | 95.58% | 94.03% | 95.00% | 94.84% | **95.93%** |
| AK | 89.71% | 83.07% | 89.24% | **92.01%** | 90.53% |
| BKL | 90.72% | 82.83% | 84.72% | 85.61% | **85.76%** |
| DF | 99.07% | 88.05% | 84.88% | 91.75% | **92.93%** |
| VASC | 100% | 93.70% | 92.93% | 96.33% | 95.28% |
| SCC | 90.87% | 79.11% | 85.33% | 86.87% | **88.75%** |

Thus, to prove the function of RL-c, firstly, we compare Pure MT and SCMT in which RL-c is additionally incorporated, on both the Pneumonia Chest X-ray dataset and ISIC challenge 2019 dataset, and the experimental results are shown in Table V and Table VI respectively. As illustrated in Table V, compared with MT, with RL-c, SCMT improves the Accuracy and F1 value by 1.01% and 1.39%, respectively. For the average AUC in Table VI, SCMT achieves a higher score (90.46%) than pure MT (88.60%). Secondly, we further visualize the inter-sample similarity matrices between the on-the-fly samples and reference samples in $RSP$. The blue blocks in ground truth stand for intra-class similarities (i.e., similarities between samples from the same categories). It can be seen that compared with the Pure MT framework (Fig. 3 (a)), the similarity matrix presented by our SCMT network (Fig. 3 (b)) is more reasonable with higher similarities between samples of the same class (i.e., more similar to the similarity metrics of the ground truth).

**Ablation Studies on SSMT Network**. The SSMT network is proposed to solve Problem $\mathbb{B}$ discussed in the Introduction Sector, i.e., the overly dense sample distribution that may affect the derivation of effective relative location features. Specifically, SSMT involves contrastive learning to push views across samples away, thus, sparsifying the sample distribution. To verify the contribution of SSMT, firstly we compare SCMT with DT-SC (SCMT+SSMT) on the ISIC challenge 2019 dataset and Pneumonia Chest X-ray dataset and summarize the experimental results in Table V and Table

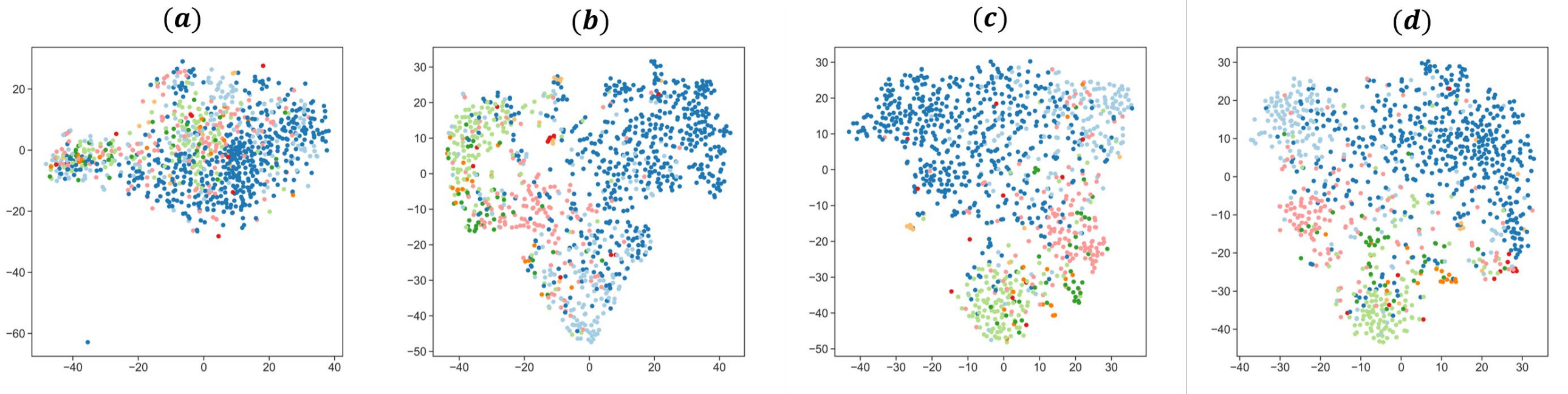

**Fig. 4**. 2D visualization of sample distribution after 20-spoch of sample scatter phase: (a) without any scatter, (b) with only supervised methods to perform sample scatter, (c) with SSMT network to perform sample scatter with $\lambda_{CL}$ set to 0.05, (d) with SSMT network to perform sample scatter with $\lambda_{CL}$ set to 0.1.

TABLE VII
QUANTITATIVE RESULTS OF DT-SC AND SRC-MT UNDER DIFFERENT $N_l$ AND $N_u$ SETTINGS.

| Method | $N_l$ | $N_u$ | Accuracy | AUC | Precision | Recall | F1 |
|---|---|---|---|---|---|---|---|
| Upper Bound | 100% | 0% | 92.31% | 97.82% | 91.63% | 91.63% | 91.55% |
| Baseline | 5% | 0% | 81.57% | 92.22% | 80.21% | 81.04% | 80.30% |
| SRC-MT [7] | 5% | 95% | 82.58% | 92.52% | 81.42% | 82.28% | 81.58% |
| **DT-SC** | **5%** | **95%** | **83.33%** | **94.50%** | **82.36%** | **82.36%** | **82.19%** |
| Baseline | 10% | 0% | 82.53% | 92.65% | 81.11% | 82.31% | 81.33% |
| SRC-MT [7] | 10% | 90% | 85.90% | 95.74% | 85.57% | 85.25% | 84.98% |
| **DT-SC** | **10%** | **90%** | **87.50%** | **96.10%** | **86.92%** | **86.53%** | **86.54%** |
| Baseline | 15% | 0% | 85.74% | 94.86% | 83.74% | 85.07% | 84.07% |
| SRC-MT [7] | 15% | 85% | 87.18% | 95.26% | 86.35% | 86.17% | 85.98% |
| **DT-SC** | **15%** | **85%** | **87.82%** | **96.11%** | **86.85%** | **87.65%** | **86.85%** |
| Baseline | 20% | 0% | 87.66% | 96.46% | 86.58% | 86.97% | 86.59% |
| SRC-MT [7] | 20% | 80% | 89.10% | 97.66% | 88.02% | 88.55% | 88.05% |
| **DT-SC** | **20%** | **80%** | **90.58%** | **97.74%** | **89.73%** | **89.87%** | **89.80%** |

VI. On the one hand, according to Table V, the proposed DT-SC (SSMT+SCMT) framework improves SCMT with 0.10% higher in average AUC, proving the effectiveness of the SSMT network. On the other hand, according to Table VI, with SSMT, DT-SC achieves a top-level performance on the Pneumonia Chest X-ray dataset. Secondly, besides the quantitative results, we give a qualitative comparison result, which visualizes the sample distribution in the sample space to prove that SSMT does perform sample-level scatter. Specifically, it can be seen in Fig. 4 that the sample distribution after the Sample Scatter Phase with 20-epoch (Fig.4 (c)), is much more scattered with our SSMT network involved, compared with the supervised method (Fig. 4 (b)) which only utilize the labeled data during sample scatter phase. Moreover, with $\lambda_{CL}$ in (10) increasing, the sample distribution grows sparser, further proving that contrastive learning does help to scatter sample distribution.

**Ablation Studies on the different $N_l, N_u$ settings.** Besides, to verify the effectiveness of our method, we further study the influence of different proportionality between the amount of labeled data $N_l$ and unlabeled data $N_u$. As seen in Table VII, our DT-SC achieves consistent improvement regardless of $N_l, N_u$ settings, compared with the baseline which only utilizes labeled data. Moreover, we further compare our DT-SC with the SOTA method SRC-MT, both of which benefit from extracting inter-sample relation information. However, due to the elaborate adoption of the Ac-L and RL-c paradigms, which explore both absolute and relative location relationships between samples, our DT-SC is capable of extracting more semantic information from unlabeled data. Therefore, regardless of the settings for $N_l$, $N_u$, our DT-SC consistently outperforms SRC-MT across all evaluation metrics, thereby demonstrating the superiority of our method.

**Ablation Studies on the Asymmetric Structure.** To prevent the student model from collapsing to the sample scatter teacher model in SSMT when performing contrastive learning, we extend SSMT with an asymmetric structure. To evaluate the effectiveness of the asymmetrical structure of SSMT, we compare the proposed DT-SC with a symmetrical DT-SC without the MLP predictor $Pred$ in (9). As observed from Fig. 5, the asymmetrical structure improves Accuracy, AUC, and F1 value by 1.43%, 1.48%, and 1.62% respectively. These improvements demonstrate the asymmetrical structure's crucial role in enabling the student model to learn independently and generalize effectively, rather than merely imitating the teacher model [22].

**Ablation Studies on different hyper-parameter settings.** To validate the learning strategy in (10) and (11), we conduct extensive experiments with the different values of hyper-parameters on the Pneumonia Chest X-ray dataset. Initially, we set $\lambda_A$ to 1.0 and varies $\lambda_R$ from 0.0 to 1.5. As depicted in Fig. 6 (a), the performance of our method consistently improves with the increase of $\lambda_R$, which has a direct impact on the RL-c paradigm, and hits its peak when $\lambda_R$ is set to 1.0. However, when $\lambda_R$ increases to 1.5, the model's performance begins to decline, possibly due to an excessive focus on relative location relationships, leading to overfitting. Subsequently, we fix $\lambda_R$ to 1.0 and vary $\lambda_A$ from 0.0 to 1.5. As shown in Fig. 6 (b), the trend for $\lambda_A$ mirrors that of $\lambda_R$ with DT-SC achieving peak performance when $\lambda_A$ is set to 1.0. Meanwhile, the performance of DT-SC is more sensitive to changes in $\lambda_A$ than $\lambda_R$. This relatively heightened sensitivity may be due to the predominant role of Ac-L over Rc-L. Therefore, we set both $\lambda_A$, $\lambda_R$ to 1.0. Then, we conduct other trial experiments to find the most suitable $\lambda_{CL}$ to balance the contribution of supervised loss and contrastive loss during the sample scatter phase. Specifically, we first fix $\lambda_A$, $\lambda_R$ to 1.0, and set $\lambda_{CL}$ from a relatively small range from 0.0 to 0.15. The results of these trials, as summarized in Fig. 6 (c), indicate that our method is not sensitive to $\lambda_{CL}$ . Furthermore, while contrastive loss is crucial for maintaining a dispersed feature space to facilitate the exploration of more effective feature representations, excessive dispersion can lead to an improper feature distribution, adversely affecting the model. Therefore, we set to 0.1 to achieve a comparatively superior performance.

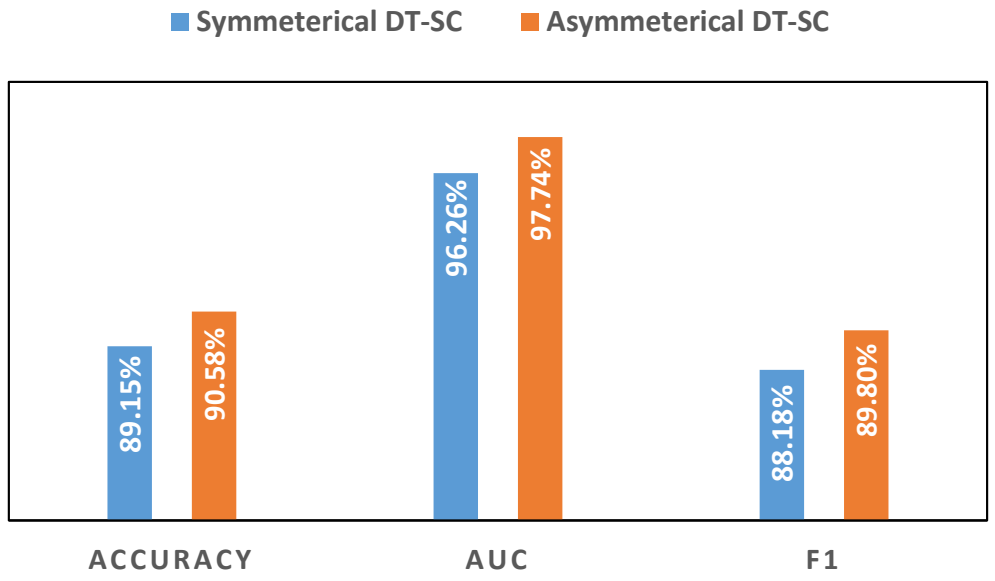

**Fig. 5.** Evaluation of asymmetric structure on Chest X-ray dataset.

TABLE VIII
ABLATION STUDIES OF MEMORY SIZE ON PNEUMONIA CHEST XRAY DATASET.

| Mem. Size | Accuracy | Mem. Occ | FLOPs |
|---|---|---|---|
| 2048 | 89.99% | 16.43G | 8.0B |
| 4096 | 90.58% | 18.69G | 7.4B |
| 8192 | 90.21% | 23.78G | 6.8B |
| 16,384 | -- | out-of-memory | -- |

**Ablation Studies on the *RSP* and $\mathcal{M}$ size.** According to Problem $\mathbb{A}$ discussed in the Introduction Sector, too few reference samples may hinder the derivation of accurate relative location features, and assuring *RSP* to store enough samples could address this problem. Herein, we conduct experiments on *RSP* size to further prove this assumption. Specifically, we progressively enlarge *RSP* until its size reaches the dimension of absolute location features ($K = 1024$). According to Fig. 6(d), the accuracy rises as *RSP* gets larger, and when *RSP* size reaches 1024, DT-SC achieves top-level performance. Besides, to better perform contrastive learning in SSMT, it is crucial to store enough key features in the memory bank $\mathcal{M}$. Thus, we further analyze the impact of the size M for the memory bank $\mathcal{M}$ on the performance (e.g., accuracy), training efficiency (e.g., floating-point operation per second, FLOPs), and memory occupancy (Mem. Occ.) on the pneumonia chest x-ray dataset. As shown in Table VIII, increasing M from 2048 to 4096 significantly improves accuracy from 89.99% to 90.54%, while memory occupancy rises from 16.43GB to 18.69GB and computational demand decreases slightly, with FLOPs dropping from 8.0B to 7.4B. However, further increasing M to 8192 results in memory occupancy reaching near its upper limit at 23.78GB, and a decrease in accuracy to 90.21%. This indicates a performance plateau as the size of key features increase will not yield significant improvements. Consequently, we set M to 4096 for an optimal balance between performance and memory usage.

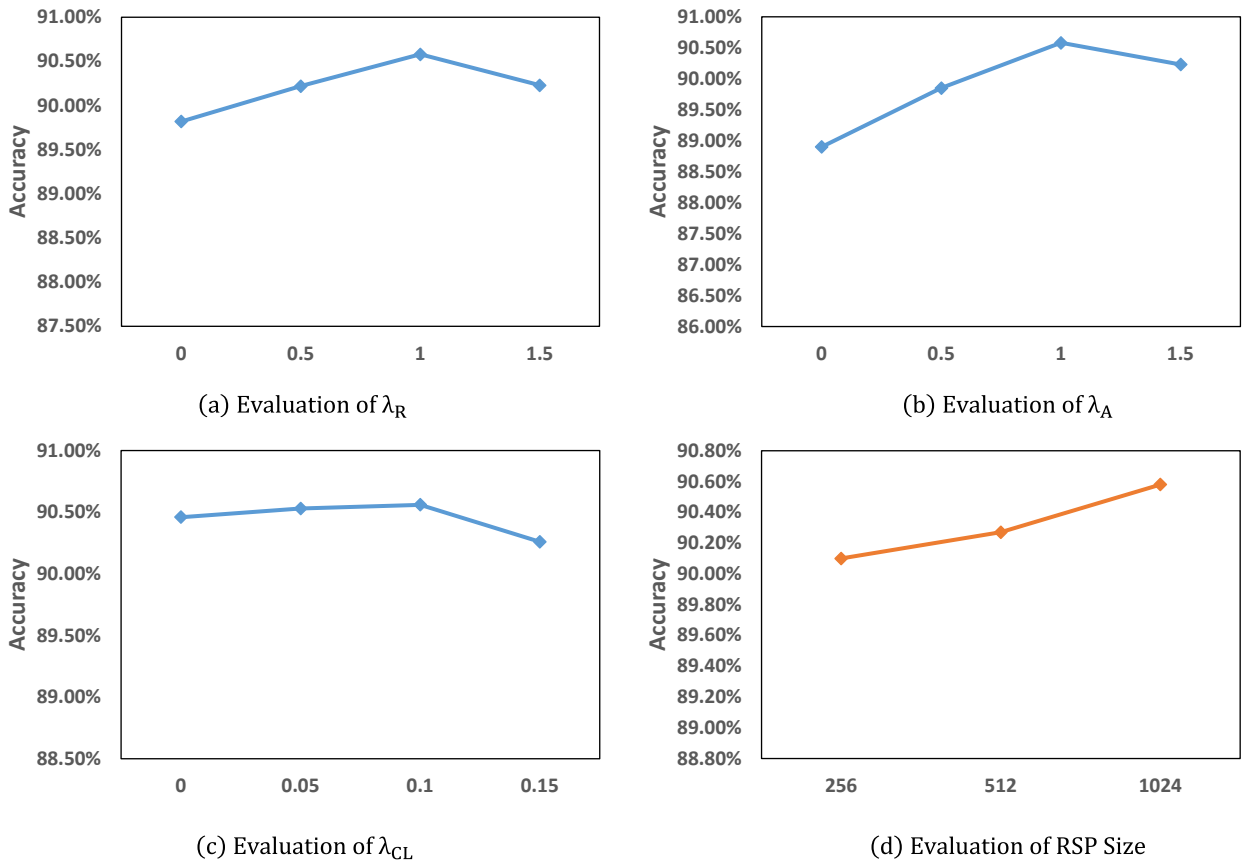

**Fig. 6.** Evaluation of key parameters, $\lambda_R$(a), $\lambda_A$(b), $\lambda_{CL}$(c) and RSP size(d) on the Pneumonia Chest X-ray dataset.

## V. DISCUSSION

In this paper, we focus on semi-supervised medical image classification, which alleviates the annotation of labeled data and utilizes numerous unlabeled data to achieve higher classification results. The key insight of our DT-SC framework is to fully exploit hidden information inside the relative differences among different samples. Specifically, in addition to absolute location consistency (AL-c) in Mean Teacher (MT), we propose a novel relative location consistency (RL-c) paradigm to impose consensus upon relative locations, thus fully extracting semantic information from the relative differences across samples. This multi-perspective consistency regularization addresses the limitations of traditional single-view learning, thereby enhancing the robustness of the semi-supervised medical image classification method against unlabeled data. Furthermore, we design a sample consistency mean teacher (SCMT) network to implement both AL-c and RL-c. Then, to overcome the problem of overly dense sample distribution involved by RL-c, we design a sample scatter mean teacher (SSMT) network to perform sample scatter in a contrastive learning manner, aiming to address the common overly dense sample distribution issue in medical imaging scenarios. Notably, our RL-c differs fundamentally from the nearest-neighbor clustering strategy. The nearest-neighbor clustering strategy classifies a sample based on the majority category of its nearest neighbors, which focuses on distance metrics. In contrast, RL-c is concerned with the consistency of relative feature positioning across different views of the same sample with reference samples. This focus is not on classification or distance metrics but on achieving a consensus in relative locations, which is more aligned with the diagnostic reasoning of medical practitioners.

Despite the considerable progress we made on the 2D class-balanced dataset (i.e., Pneumonia Chest X-ray dataset), 2D class-imbalanced dataset (i.e., ISIC Challenge 2019 dataset), and 3D MRI dataset (i.e., Alzheimer's dataset), there are still some limitations. Firstly, the efficacy of the RL-c paradigm is still inadequate. On the one hand, the selection strategy of reference samples can be improved. In the future, we will try to design a more efficient strategy to select more reasonable reference samples. On the other hand, the calculation of similarities across samples is quite limited, lacking further exploration. Secondly, apart from the classification tasks, dense prediction tasks (e.g., semantic segmentation) also play significant roles in the medical image analysis field. How to refactor the framework to adapt to such dense prediction tasks could be another challenge, which will also be an important issue we would like to focus on in the future.

## VI. Conclusion

In this paper, we present a novel Dual Teacher Sample Consistency framework consisting of a Sample Consistency Mean Teacher (SCMT) network and a Sample Scatter Mean Teacher (SSMT) network. On the one hand, the SCMT network encourages both RL-c and AL-c to explore the semantic information inside the unlabeled data. On the other hand, the SSMT network performs sample-level scatter in a contrastive learning manner to solve the problem of overly dense sample distribution and assure the robustness of relative location features. Extensive experiments demonstrate the superior performance of our DT-SC over state-of-the-art semi-supervised learning methods.